\documentclass[10pt,twocolumn,letterpaper]{article}

\usepackage{cvpr}
\usepackage{times}
\usepackage{epsfig}
\usepackage{graphicx}
\usepackage{amsmath}
\usepackage{amssymb}
\usepackage{array}
\usepackage{multirow}
\usepackage{caption}
\usepackage{subcaption}
\usepackage{float}
\usepackage{breqn}
\usepackage{booktabs}
\usepackage{multicol}
\usepackage{dblfloatfix}



\usepackage[inline]{enumitem}

\DeclareMathOperator*{\argmax}{argmax}

\usepackage[pagebackref=true,breaklinks=true,letterpaper=true,colorlinks,bookmarks=false]{hyperref}

\cvprfinalcopy % *** Uncomment this line for the final submission

\def\cvprPaperID{5557} % *** Enter the CVPR Paper ID here

\ifcvprfinal\fi
\begin{document}

\definecolor{colorabd}{RGB}{220,100,0}
\newcommand{\abd}[1]{\textcolor{colorabd}{#1}}

%%%%%%%%% TITLE
\title{PandaNet : Anchor-Based Single-Shot Multi-Person 3D Pose Estimation}

% \begin{figure}[H]
% \onecolumn\includegraphics[width=\textwidth]{viz_jta.png}
% \end{figure}

%  \begin{multicols}{2}

\author{Abdallah Benzine$^{\star,\dagger}$, Florian Chabot$^{\star}$, Bertrand Luvison$^{\star}$, Quoc Cuong Pham$^{\star}$, Catherine Achard $^{\dagger}$\\
$^{\star}$ CEA LIST Vision and Learning Lab for Scene Analysis \\
$^{\dagger}$ Sorbonne University, CNRS, Institute for Intelligent Systems and Robotics\\
%{\tt\small abdallah.benzine@cea.fr}
}
% \maketitle

%\thispagestyle{empty}

\twocolumn[{%
\renewcommand\twocolumn[1][]{#1}%
\maketitle
\begin{center}
    % \centering
    % \includegraphics[width=0.96\textwidth]{viz_principale_5.png}
        \includegraphics[width=0.75\textwidth]{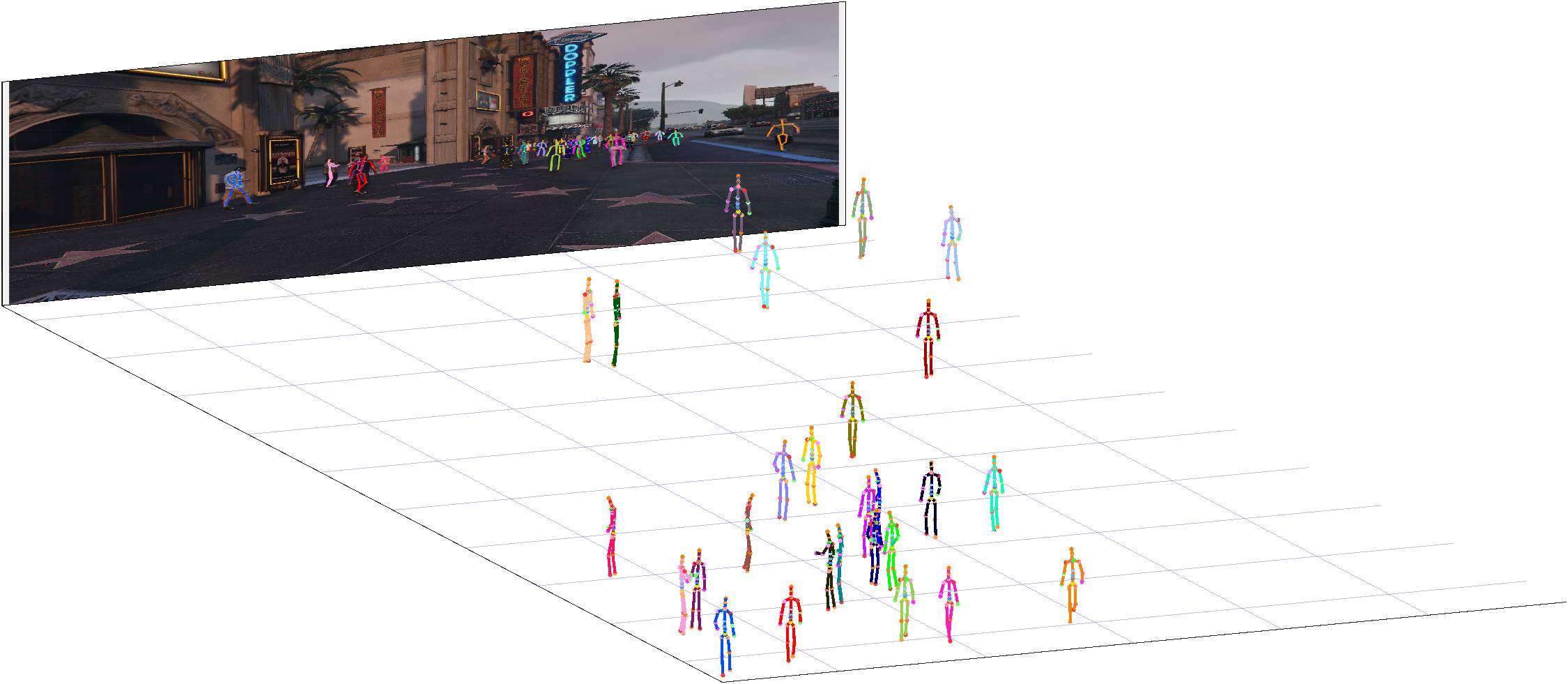}

    \captionof{figure}{Qualitative results of  PandaNet on JTA dataset \cite{fabbri2018learning} which consists in images with many people (up to 60), a large proportion of people at low resolution and many occlusion situations. Most of the previous 3D human pose estimation studies mainly focused on the single-person case or estimate 3D pose of few people at high resolution. In this paper, we propose an anchor-based and single-shot multi-person 3D pose estimation framework that allows the pose estimation of a large number of people at low resolution. Ground-truth translations and scales are used for visualisation.}
    \label{fig:predsjta}
 \end{center}%
}]

% \begin{tablehere}
% \includegraphics[width=\textwidth]{viz_jta.png}
% \end{tablehere}

%%%%%%%%% ABSTRACT
\begin{abstract}
 Recently, several deep learning models have been proposed for 3D human pose estimation. Nevertheless, most of these approaches only focus on the single-person case or estimate 3D pose of a few people at high resolution. Furthermore, many applications such as autonomous driving or crowd analysis require pose estimation of a large number of people possibly at low-resolution. In this work, we present PandaNet (Pose estimAtioN and Dectection Anchor-based Network), a new single-shot, anchor-based and multi-person 3D pose estimation approach. The proposed model performs bounding box detection and, for each detected person, 2D and 3D pose regression into a single forward pass. It does not need any post-processing to regroup joints since the network predicts a full 3D pose for each bounding box and allows the pose estimation of a possibly large number of people at low resolution.  
 To manage people overlapping, we introduce  a Pose-Aware Anchor Selection strategy. 
Moreover, as imbalance exists between different people sizes in the image, and joints coordinates have different uncertainties depending on these sizes, we propose a method to automatically optimize weights associated to different people scales and joints for efficient training. 
 PandaNet surpasses previous single-shot methods on several challenging datasets: a multi-person urban virtual but very realistic dataset (JTA Dataset), and two real world 3D multi-person datasets (CMU Panoptic and MuPoTS-3D). 
\end{abstract}

%%%%%%%%% BODY TEXT
\section{Introduction}

\begin{figure}[t]
    \centering
    \begin{subfigure}[b]{0.22\linewidth} % "0.45" donne ici la largeur de l'image
        \centering \includegraphics[width=\textwidth]{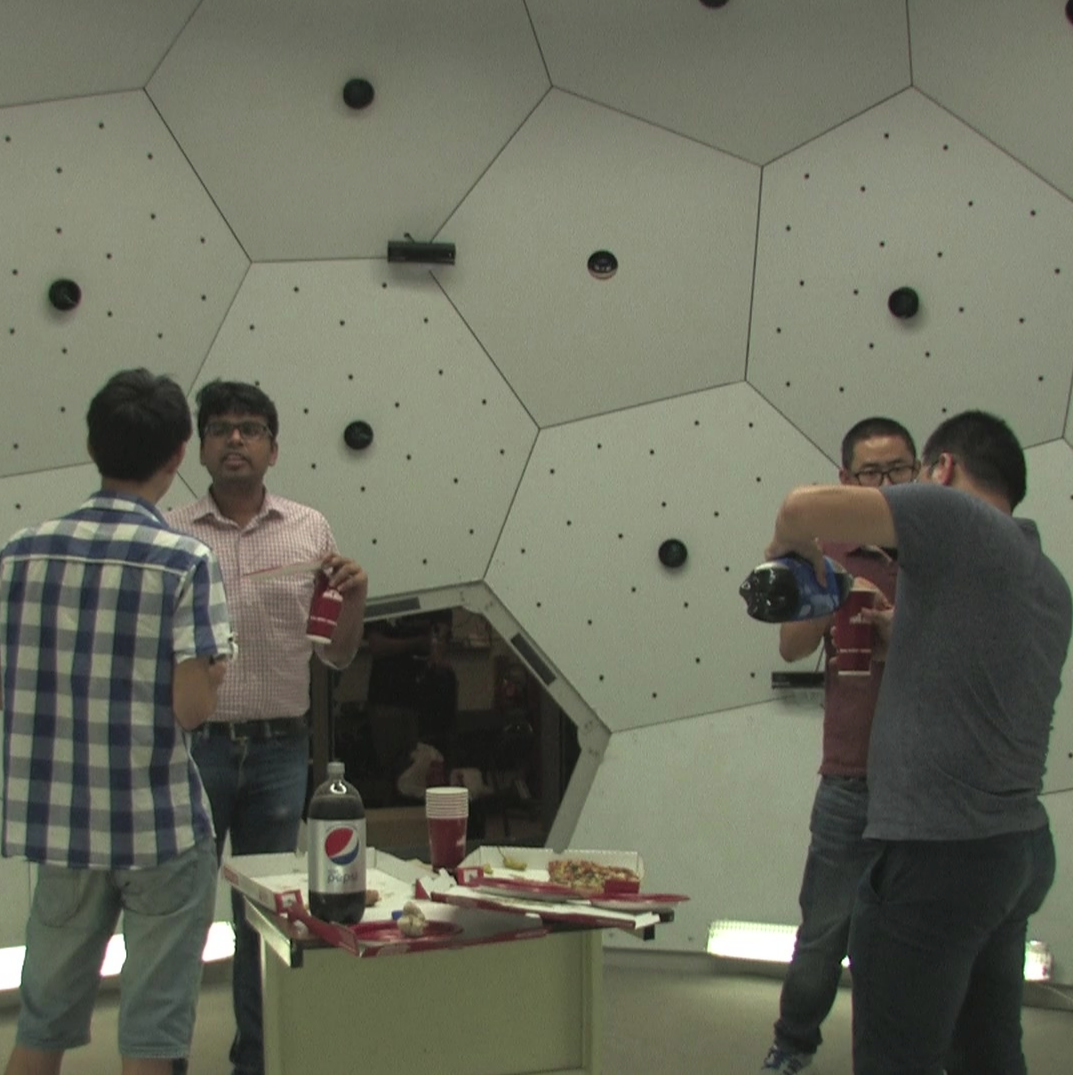}
        \caption{}\label{fig:3dcont_panoptic}
    \end{subfigure}
    ~ % ce symbole ajoute un espacement horisontal entre les premières deux images
    \begin{subfigure}[b]{0.22\linewidth}
        \centering \includegraphics[width=\textwidth]{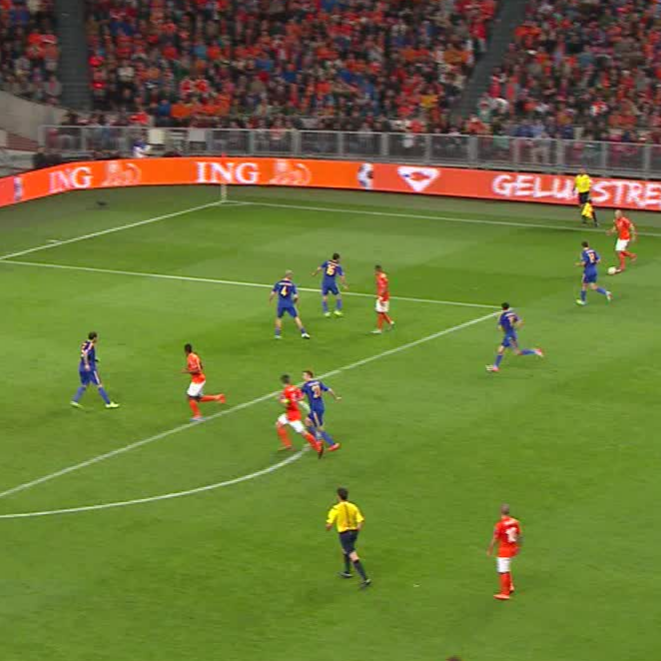}
        \caption{}\label{fig:3dcont_soccer}

    \end{subfigure}
        ~ % ce symbole ajoute un espacement horisontal entre les premières deux images
    \begin{subfigure}[b]{0.22\linewidth}
        \centering \includegraphics[width=\textwidth]{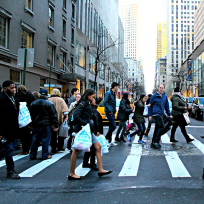}
        \caption{}\label{fig:3dcont_pedestrian}
    \end{subfigure}
            ~ % ce symbole ajoute un espacement horisontal entre les premières deux images
    \begin{subfigure}[b]{0.22\linewidth}
        \centering \includegraphics[width=\textwidth]{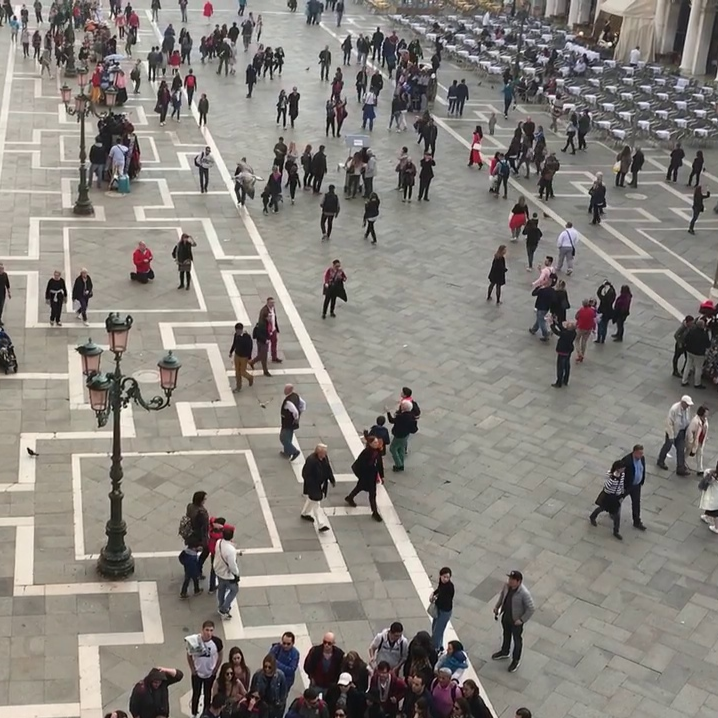}
        \caption{}\label{fig:3dcont_venise}
    \end{subfigure}
\caption{\textbf{Different real-world contexts for 3D human pose estimation}. Recent 3D multi-person estimation approaches focus on 3D pose estimation of a few people close to the camera like in a). This context is yet challenging because of frequent inter-people occlusions. 3D pose estimation is even more difficult in applications such as sport analysis (b), autonomous driving (c) or crowd analysis (d) with a large number of people at low resolution.}
\label{fig:3d_contexts}

\end{figure}

3D human pose estimation is a common addressed problem in Computer Vision. It has many applications such as crowd analysis, autonomous driving, human computer interaction or motion capture.  3D pose is a low dimensional and interpretable representation that allows to understand and anticipate human behavior. 
% In autonomous driving context, the sooner the 3D human pose of a pedestrian is estimated, the more likely the vehicle will be able to anticipate his behaviour.  
Great progresses have been achieved thanks to large scale datasets with 2D annotations (LSP \cite{Johnson10}, MPII \cite{andriluka14cvpr}, COCO \cite{lin2014microsoft}, CrowdPose \cite{li2018crowdpose}) and 3D annotations (Human 3.6M \cite{ionescu2014human3}, MPI-INF-3DHP \cite{mono-3dhp2017}, MuCo-3D-HP \cite{mehta2017single}, CMU Panoptic \cite{Joo_2017_TPAMI}). 
Nevertheless, this problem remains hard as the human body is an articulated object whose terminal joints are very mobile and thus difficult to be precisely located.  In addition,  real-world applications require to handle a large number of people and crowded images like the ones in Figure \ref{fig:3dcont_soccer}, \ref{fig:3dcont_pedestrian} and \ref{fig:3dcont_venise}. To handle these challenging conditions, models need to be robust to people occlusions and to low resolution (\textit{i.e} people that occupy a small portion of the image). They also need to be fast and to handle a large number of people. Most existing approaches focus on 3D pose estimation of either a single person or a limited number of people that are relatively close to the camera.

% To the best of our knowledge, no prior work has addressed the problem of 3D human pose estimation on urban and crowded scenes.  

 Although top-down and two-stage based methods are currently considered as best performing in the state of the art, these approaches become slow in crowded scenes as their computation complexity increases with the number of people. On the contrary, bottom-up approaches perform their forward pass with a constant complexity. Existing bottom-up single-shot methods rely on heatmaps prediction followed by complex post-processing steps to properly regroup joints detections into full human skeletons. 3D coordinates of joints are stored in maps at their corresponding 2D position. Consequently, if 2D localisation or 2D association of joints fails, 3D pose estimation will also fail. These approaches can also fail for other reasons. First, they lack precision at low resolution because of the downsampling factor between the input image and the predicted heatmaps.  Second, heatmaps are usually not sharp enough to distinguish two very close joints of the same type.  Finally, 3D coordinates of two overlapping joints cannot be stored at the same 2D location causing erroneous 3D pose estimation. For all these reasons, we believe that heatmap based approaches are not suited for robust 3D pose estimation for occluded people at low resolution.   
 
% We argue that the main reason why single shot 3D human pose approaches fail is because they are heatmap based and have not used any mechanism to deal with different person scales that occur in real world crowded images. 
%  Predicted heatmaps in these approaches are usually smaller by a scaling factor than the original image while 2D joint detection needs high resolution heatmaps.
% It follows that heatmap based methods fail to detect low and very low resolution people that are frequent in real world crowded environments.

% We argue that heatmaps are not necessary to perform single-shot multi-person 3D human pose estimation. 

In this paper, we introduce PandaNet (Pose estimAtioN and Dectection Anchor-based Network), a new single-shot approach that performs bounding box detection in a dense way and regresses 2D and 3D human poses for each detected person. To this end, three contributions are proposed.

% This method has a constant complexity regardless of the analysed scene and is subject-centric rather than joints-centric. 

First, an anchor based representation is adopted. An anchor that matches a subject stores its full 3D pose. This avoids problems induced by occlusion of joints. Additionally,  this anchor-based formulation allows lower resolution outputs than heatmap one since a single output pixel is enough to store the entire subject's pose. This property is important to efficiently process people at low resolution.
% Indeed, 2D detection of joints in this context is hard for multi-person heatmap based methods. For instance,  a 20 pixels height subject in the original image has a height in the heatmap of only 5 pixels (supposing a downsampling factor of 4). Thus, the prediction of multiple 2D peaks corresponding to the person joints is very hard. Moreover, heatmap based methods need an association step to group separate joint predictions into full skeletons.  People at low resolution and strong occlusions make this association even more difficult than joint detection. 
%Second, contrary to previous single-shot multi-person 3D estimation methods, the proposed anchors based approach is designed to handle different people sizes inside the same image and in particular people at low resolution. 

Second, a Pose-Aware Anchor Selection strategy discards ambiguous anchors during inference. Indeed, ambiguous anchors overlap parts of several people and do not allow a readout of consistent 3D poses.

Third, an automatic weighting of losses with homoscedastic uncertainty handles imbalance between people sizes in the image and uncertainties associated to human pose predictions.

Contrary to previous top-down multi-person approaches, PandaNet has a forward inference complexity that does not depend on the number of people in the image. It can efficiently process images with a large number of people (cf. Figure \ref{fig:predsjta}). 
The proposed model is validated on three 3D datasets. The first one is the Joint Track Auto dataset (JTA) \cite{fabbri2018learning}, a rich, synthetic but very realistic urban dataset with a large number of people (up to 60) and occlusion situations. The second one is the Panoptic dataset \cite{Joo_2017_TPAMI}, an indoor dataset with many interactions and social activities. The third one is the MuPoTS-3D dataset \cite{mehta2017single}, a dataset with a reduced number of people but various indoor and outdoor contexts. Our results outperform those of previous single-shot methods on all these datasets. 

% Our contributions can be summarized as follows : 
% \begin{itemize}
%     \item We propose a new single shot anchor-based framework to estimate multi-person 3D poses from a single RGB image
%     \item Our model  detects each human in the image, estimates their 2D pose and predicts the root relative 3D coordinates of their joints.
%     \item An Intersection Over Union 2D pose loss is proposed 
%     \item A Pose-Aware weighting strategy is proposed to train the classification head of our network
%     \item A training loss is proposed to automatically weight different human scales imbalance and uncertainties associated to different joints predictions. 
%     \item We show that our method significantly outperform previous multi-person 3D methods on virtual and real-world benchmarks. 
% \end{itemize}

\section{Related Work}

\textbf{2D multi-person pose estimation.}
 Two main approaches in multi-person 2D pose estimation cas be distinguished: top-down and bottom-up approaches. Methods of the former category first perform human detection then estimate a 2D pose within the detected bounding box. On the other hand, bottom-up approaches localise all human body keypoints in an input image and then group these keypoints into full 2D human skeletons.  Each of them has advantages and disadvantages. Bottom-up methods are generally faster and seems more suited for crowded scenes since they process the entire image at once. However, available benchmarks show that top-down approaches are more accurate as they  process all subjects at the same scale.

State of the art bottom-up approaches \cite{cao2016realtime, newell2017associative, kocabas2018multiposenet} differ on their association method. Cao \etal \cite{cao2016realtime} propose Part Affinity Fields that are 2D vectors modeling the associations between children and parent joints. They are used to regroup  2D predictions of joints into full 2D skeletons.  Newell \etal \cite{newell2017associative} perform this grouping by training the network to predict similar tag values to joints belonging to the same person and different tag values for joints belonging to different people. Kreiss \textit{el al.} \cite{kreiss2019pifpaf} propose a bottom-up approach to handle  people at low resolution in crowded images.  Their model predicts Part Intensity Fields (PIF) that are similar to the offsets and heatmaps  in \cite{papandreou2017towards} and Part Associative Field (PAF) that has a composite structure.

 The methods in \cite{he2017mask, chen2018cascaded, papandreou2017towards, xiao2018simple, li2019crowdpose, golda2019crowdposeestimation} are top-down approaches.  Mask R-CNN \cite{he2017mask} detect keypoints as a segmentation task.  The method in \cite{papandreou2017towards} performs 2D offsets and 2D heatmaps prediction and fuses these predictions to generate more precise heatmaps. Chen \etal \cite{chen2018cascaded} propose a cascaded pyramid network to generate 2D poses with a refinement process that focuses on hard keypoints. Xiao \etal \cite{xiao2018simple} present a simple architecture with deep backbone and several upsampling layers. 

While top-down approaches achieve higher 2D pose estimation scores in standard benchmarks than bottom-up approaches, most of these approaches fail in scenes with frequent and strong occlusions. Indeed, these methods depend on the predicted bounding boxes. In crowded scenes, bounding boxes, even if correct, may contain parts of other people. This situation is not well managed by existing methods. Li \etal \cite{li2019crowdpose} introduce a new benchmark to evaluate 2D human pose models on crowded scenes and a method that performs multi-peak predictions for each joint and a global maximum joint association. Golda \etal \cite{golda2019crowdposeestimation} propose an approach that explicitly detects occluded body parts, uses a data augmentation method to generate occlusions and exploits a synthetic generated dataset.

\textbf{Single-person 3D pose estimation.} There are two categories of single person 3D pose estimation approaches: direct and reconstruction approaches. Direct approaches estimate the 3D pose directly from an input image while reconstruction methods first take as input 2D poses provided by a 2D pose estimation model and lift them to the 3D space.

The approaches described in \cite{fang2018learning, martinez2017simple} are reconstruction-based methods.  Martinez \textit{et al.} \cite{martinez2017simple} regress 3D pose from 2D pose input by using a simple architecture with residual connections and batch normalisation. Fang \etal \cite{fang2018learning} use a pose grammar model that takes into account the connections between human joints. These reconstruction approaches are limited by the 2D pose estimator performance and do not take into account important images clues, such as contextual information, to make the prediction.

The models in  \cite{li20143d, pavlakos2017coarse, sun2017compositional, sun2018integral, tekin2016structured, zhou2017towards, yang20183d} are direct approaches. Li \textit{et al.} \cite{li20143d} simultaneously learn 3D pose regression and body part detection. Tekin \textit{et al.} \cite{tekin2016structured} predict 3D poses in an embedding space learned by an autoencoder. Pavlakos \textit{et al.} \cite{park20163d} adopt a volumetric representation and a coarse to fine architecture to predict 3D poses. Sun \textit{et al.} \cite{sun2017compositional} take into account the connection structure between joints by proposing a compositional loss. Sun \textit{et al.} \cite{sun2018integral} use the soft-argmax layer to extract 3D coordinates from a 3D volumetric representation in a differentiable way. Zhou \textit{et al.} \cite{zhou2017towards} use a geometric loss based on bones constraints to weakly supervise the depth regression module on \textit{in the wild} images. Yang \textit{et al.} \cite{yang20183d} improve generalisation to \textit{in the wild} images thanks to an adversarial loss.

\textbf{Multi-person 3D pose estimation.} Multi-person 3D pose estimation has been less studied. It is a difficult problem that adds to the 2D multi-person management difficulty, that of depth estimation.
Zanfir \textit{et al.} \cite{zanfir2018deep} estimate the 3D human shape from sequences of frames. A pipeline process is followed by a 3D pose refinement based on a non-linear optimisation process and semantic constraints. In a top-down approach, Rogez \textit{et al.} \cite{rogez2017lcr, rogez2019lcr} generate human pose proposals that are classified into anchor-poses and further refined using a regressor. Moon \textit{et al.} \cite{moon2019camera} propose a camera distance aware multi-person top-down approach that performs human detection (DetectNet), absolute 3D human localisation (RootNet) and root relative 3D human pose estimation (PoseNet). These approaches perform redundant estimations that need to be filtered or fused, and scales badly with a large number of people.

All existing single-shot methods estimate both 2D and 3D human poses and rely on heatmaps to detect individual joints in the image.  Mehta \etal \cite{mehta2017single} propose a bottom-up approach
system that predicts Occlusion-Robust Pose Maps (ORPM) and Part Affinity Fields \cite{cao2016realtime} to manage multi-person 3D pose estimation even for occluded and cropped people. Benzine \textit{et al.} \cite{benzine2019deep, benzinepr} perform single-shot multi-person 3D pose estimation by  extending the 2D multi-person model in \cite{newell2017associative} to predict ORPM. ORPM based methods predict a fixed number of 2D heatmaps and ORPM, whatever the number of people in the image. 3D coordinates are stored multiple times in the ORPM allowing the readout of 3D coordinates at non occluded and reliable 2D positions. Nevertheless, this formulation implies potential conflicts when similar joints of different people overlap. In the same way, MubyNet \cite{zanfir2018monocular} also uses a fixed number of output maps to store 2D and 3D poses of all people in the image. However, the full 3D pose vector is stored at all 2D positions of the subject skeleton increasing the number of potential conflicts. The model learns to score the possible associations of joints to limbs and a global optimisation problem is solved to group the joints into full skeletons. XNect \cite{mehta2019xnect} estimates 3D poses in two steps. The first step improves the method of \cite{mehta2017single} by encoding only the joints' immediate context, which reduces the number of potential conflicts. The second step  refines the 3D poses.

PandaNet is a single-shot approach like \cite{zanfir2018monocular, mehta2017single, benzine2019deep, mehta2019xnect} but is anchor-based rather than heatmap-based. It is based on LapNet \cite{chabot},  a single-shot  object  detection  model which has today the best accuracy/inference time trade-off. Unlike LapNet that is
a 2D object detector, PandaNet is intended for multi-
person 3D pose estimation and differs from LapNet in the  prediction heads (introduced in subsection \ref{ssec:overview}), in the anchor selection strategy (described in subsection \ref{ss:pose-aware}) and in the automatic weighting of losses (described in subsection \ref{ss:alw}). It efficiently processes images with many people, strong occlusion and various sizes in the image with a complexity that does not depend on their number.

\section{Method}
\begin{figure*}[t]

  \centerline{\includegraphics[width=0.81\textwidth]{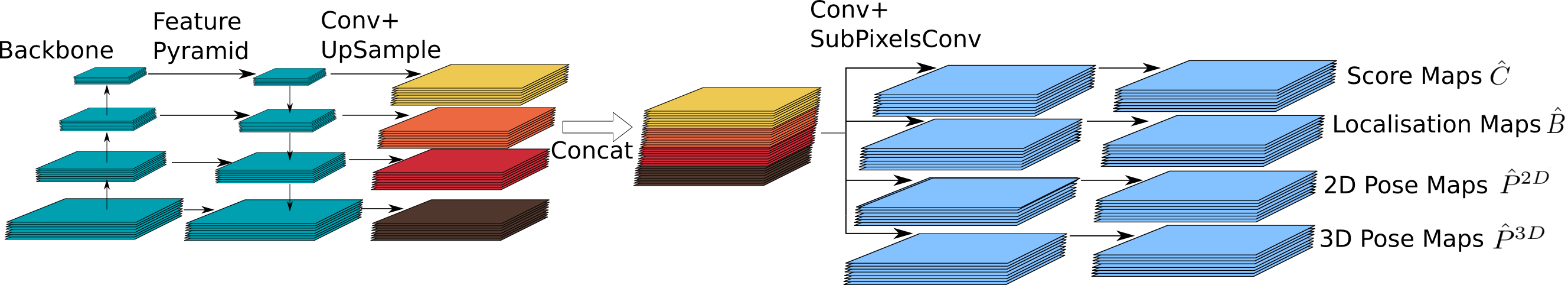}}
%  \vspace{1.5cm}
%   \centerline{(a) Model overview}\medskip

%
\caption{PandaNet architecture. The input image is passed through a backbone network. A second stage is  used to compute
pyramid feature maps at several resolutions and semantic levels (like done by FPN \cite{lin2017feature}). Four 3x3 convolutions are applied to these feature maps. The resulting maps are then upsampled to the size of the highest
resolution feature map. After multi-scale feature concatenation and subpixels convolution,  four convolutional heads are used to provide the four outputs. Each head is composed by four 3x3 convolutions and one final convolutional layer for the output.
}
\label{fig:overview}
\end{figure*}
\subsection{Overview}
\label{ssec:overview}

Given an input image, PandaNet predicts a dense set of human bounding boxes with their associated confidence  scores, 2D and 3D poses. These boxes are then filtered using non-maximum suppression to provide the final detections and human poses. 
As in most  detection approaches \cite{ren2015faster,liu2016ssd, redmon2017yolo9000, lin2017feature, chabot}, our model uses predefined anchors. These anchors are computed on the training dataset with the clustering method used in \cite{chabot, redmon2017yolo9000}.  We define $N_A$ to be the number of predefined human anchors used in the model,  $N_K$  the number of human joints and $H$ and $W$ the height and the width of the network output. The model returns: \begin{enumerate*}
 
    \item[(1)] Score maps $\hat{C} \in \mathbb{R}^{H \times W \times N_A}$  that contain the probability of an anchor to contain a subject
    \item[(2)] Box offsets maps  $\hat{B} \in \mathbb{R}^{H \times W  \times N_A \times 4}$
    \item[(3)] 2D joints coordinates maps  $\hat{P}^{2D} \in \mathbb{R}^{H \times W \times N_A \times N_K \times 2}$ that contain the full 2D pose vectors expressed relatively to their corresponding anchor
    \item[(4)] 3D joints coordinates maps $\hat{P}^{3D} \in \mathbb{R}^{H \times W  \times N_A \times N_K \times 3}$ that contain root relative 3D human poses.
\end{enumerate*}

PandaNet is a multi-task network based on LapNet, the single-shot object detection model proposed in \cite{chabot} which has today the best accuracy/inference time trade-off. The architecture of the proposed model, detailed in Figure \ref{fig:overview}, slightly differs from LapNet. First, sub-pixel convolutions are applied  \cite{shi2016real} to the feature maps to obtain higher resolution maps that are crucial to detect and  estimate the human pose of people at low resolution. Secondly, a 2D pose and 3D pose regression heads are added.

\subsection{ Anchor-based Multi-person Pose Formulation}

For a given image $I$, we define $\mathcal{B}=\{b_n \in \mathbb{R}^4\}$ as the set of ground truth bounding boxes $n \in [1,\ldots, N]$ and $N$ is the number of visible people. $\mathcal P^{2D}=\{p^{2D}_n \in \mathbb{R}^{2 \times N_K}\}$ and $\mathcal P^{3D}=\{p^{3D}_n \in \mathbb{R}^{3 \times N_K}\} $ are the sets of corresponding 2D and 3D human poses. 

In order to train PandaNet, a grid of anchors $A \in \mathbb{R}^{H \times W \times N_A \times 4}$  is defined. $A_{i,j,a}$ is an element of this grid at output position $(i,j)$ for anchor $a$.
Let $B \in \mathbb{R}^{H \times W \times N_A \times 4}$ be the grid of matched bounding boxes, each of its element is defined as:

\begin{equation}
B_{i,j,a} = \argmax_{b_n \in \mathcal{B}} IoU(b_n,A_{i,j,a}) 
\end{equation}

Similarly, $P^{2D}$ and $P^{3D}$ are defined respectively as the grids of matched 2D poses and 3D poses:

\begin{equation}
P^{2D}_{i,j,a} = p^{2D}_n\ |\ b_n = B_{i,j,a}
\end{equation}

\begin{equation}
P^{3D}_{i,j,a} = p^{3D}_n\ |\ b_n = B_{i,j,a}
\end{equation}

In other words, $P^{2D}_{i,j,a}$ and $P^{3D}_{i,j,a}$ are the 2D and 3D human poses of the subject matched by the anchor $A_{i,j,a}$

The Per-Object Normalised Overlap (PONO) map \cite{chabot} $O$ is used. $O_{i,j,a}$  is the IoU between anchor $A_{i,j,a}$ and ground truth $B_{i,j,a}$, normalised by the maximum overlap between $B_{i,j,a}$ and all matched anchors to this ground truth.

The positive anchors $\mathcal{A}^+$ are the set of matched anchors that have a PONO value greater than $0.5$.  Only bounding boxes and human poses associated to anchors in $\mathcal{A}^+$ will be supervised, like described in the next subsection.

\subsection{Bounding box offsets and human poses supervision}

\subsubsection{IoU based bounding-box offsets supervision}
Most detection approaches use SmoothL1 or Mean Squared Error losses to predict bounding box offsets that fit the ground truth bounding boxes. More recently, some methods prefer to optimize the Intersection over Union (IoU) loss \cite{tian2019fcos, chabot} or its extension \cite{ rezatofighi2019generalized} taking benefit of its invariance to scale.  We also used the IoU loss to supervise bounding box offsets prediction and, for an anchor $a$ at location $(i,j)$, we define the overlap function as: 
\begin{equation}
\hat{O}_{i,j,a} = IoU(B_{i,j,a}, \hat{B}_{i,j,a})
\end{equation}

where $\hat{B}_{i,j,a}$ is the predicted box, \textit{i.e} the anchor $A_{i,j,a}$ transformed with estimated offsets.  The pixel-wise localisation loss is then obtained with: 

\begin{equation}
    \mathcal{L}_{loc}({i,j,a}) = \left\{
    \begin{array}{ll}
        \left\|1-{\hat{O}}_{i,j,a} \right\|^2, & \mbox{if } A_{i,j,a} \in \mathcal{A}^+ \\
       0, & \mbox{otherwise}
    \end{array}
\right.
\end{equation}

\subsubsection{IoU based 2D human pose supervision}

While our main objective is single-shot multi-person 3D pose estimation, PandaNet also regresses 2D human poses for two reasons. Firstly, the predicted 2D poses are needed in the pose-aware pixel-wise classification loss defined in subsection \ref{ss:pose-aware}. Secondly, by minimizing the reprojection loss between a 2D human pose and a root relative 3D human pose, one can obtain the 3D human pose in the camera reference. 
Regressing 2D human poses is challenging because of large variations in scale between people. So, we introduce a IoU loss to supervise this step.
We designate by $ {P}^{2D}_{i,j,a,k}$ the ground-truth 2D coordinates in the anchor coordinate system of the joint $k$ of the subject matched with the anchor $a$.  These coordinates are obtained from the coordinates in the image space by translating them to the center of the anchor and dividing them by the width and the height of the anchor. $\hat{P}^{2D}_{i,j,a,k}$ are the corresponding predicted coordinates in the anchor space.  Two unit squares in the anchor space, $\hat{S}_{i,j,a,k}$ and $S_{i,j,a,k}$, centred at positions $\hat{P}^{2D}_{i,j,a,k}$ and $P^{2D}_{i,j,a,k}$ are defined to compute the IoU loss and the pixel-wise 2D pose loss for joint $k$: 
\begin{equation}
\hat{O}^{2D}_{i,j,a,k} = IoU(S_{i,j,a,k}, \hat{S}_{i,j,a,k})
\end{equation}

\begin{dmath}
    \mathcal{L}_{2D}({i,j,a,k}) = \left\{
    \begin{array}{ll}
        \left\|1-\hat{O}^{2D}_{i,j,a,k} \right\|^2, & \mbox{if } A_{i,j,a} \in \mathcal{A}^+ \\
       0, & \mbox{otherwise}
    \end{array}
\right.
\end{dmath}

\subsubsection{3D human pose supervision}

PandaNet is trained to predict scale normalised 3D human poses translated to the pelvis. The sum of the subject bones length is equal to 1. As all 3D poses are predicted at the same scale, an Euclidean distance is used as supervision. The pixel-wise 3D pose loss for joint $k$ between the ground-truth 3D joints coordinates $P^{3D}_{i,j,a,k}$ and their corresponding predicted coordinates $\hat{P}^{3D}_{i,j,a,k}$ is defined by : 

\begin{dmath}
    \mathcal{L}_{3D}({i,j,a,k}) = \left\{
    \begin{array}{ll}
        \left\|P^{3D}_{i,j,a,k}- \hat{P}^{3D}_{i,j,a,k} \right\|^2, & \mbox{if } A_{i,j,a} \in \mathcal{A}^+ \\
       0, & \mbox{otherwise}
    \end{array}
\right.
\end{dmath}

\begin{figure}[t]
    \centering
    \begin{subfigure}[t]{0.22\linewidth} % "0.45" donne ici la largeur de l'image
        \centering \includegraphics[width=\textwidth]{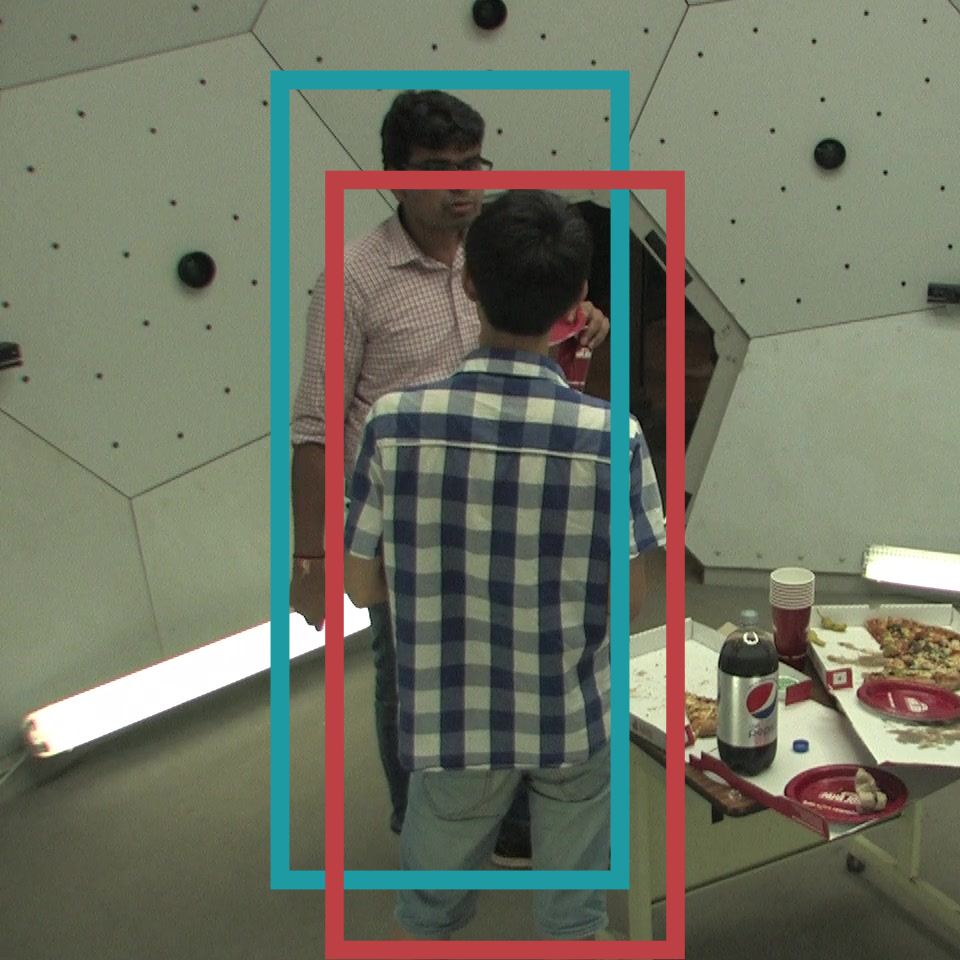}
         \caption{Input Image}\label{fig:paas_input_image}
    \end{subfigure}
    ~ % ce symbole ajoute un espacement horisontal entre les premières deux images
    \begin{subfigure}[t]{0.22\linewidth}
        \centering \includegraphics[width=\textwidth]{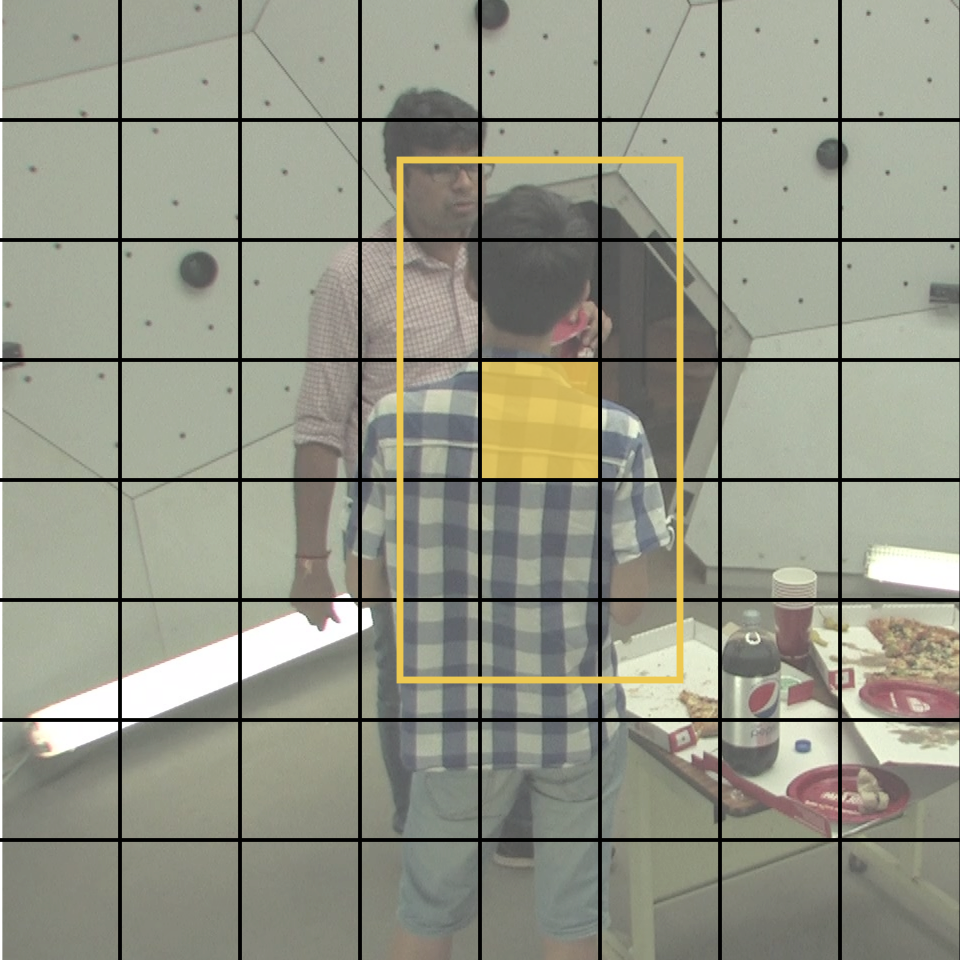}
        \caption{Grid of Anchors $\mathcal{A}$.}\label{fig:paas_A}
        % Each 2D position of the grid corresponds to an anchor like the one depicted in yellow.
    \end{subfigure}
        ~ % ce symbole ajoute un espacement horisontal entre les premières deux images
    \begin{subfigure}[t]{0.22\linewidth}
        \centering \includegraphics[width=\textwidth]{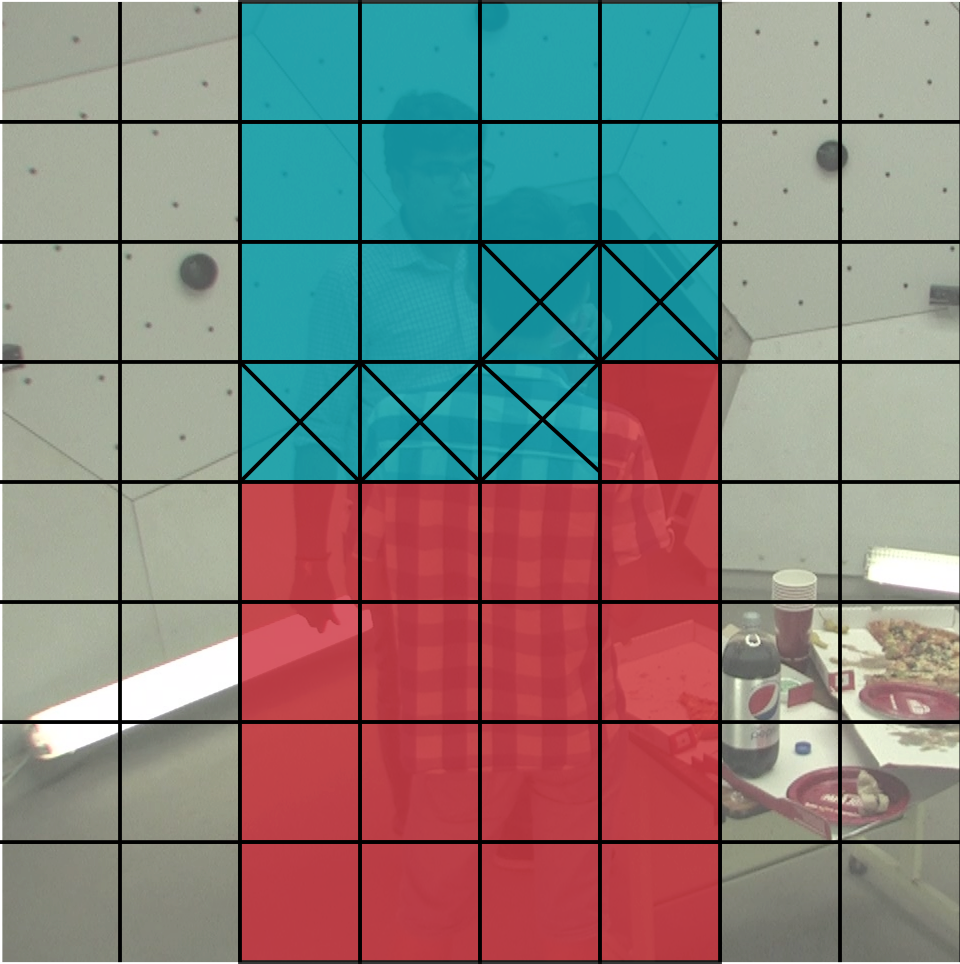}
        \caption{Matched Anchors $\mathcal{A}^+$ }\label{fig:passA+}
    \end{subfigure}
        ~ % ce symbole ajoute un espacement horisontal entre les premières deux images    
       \begin{subfigure}[t]{0.22\linewidth}
        \centering \includegraphics[width=\textwidth]{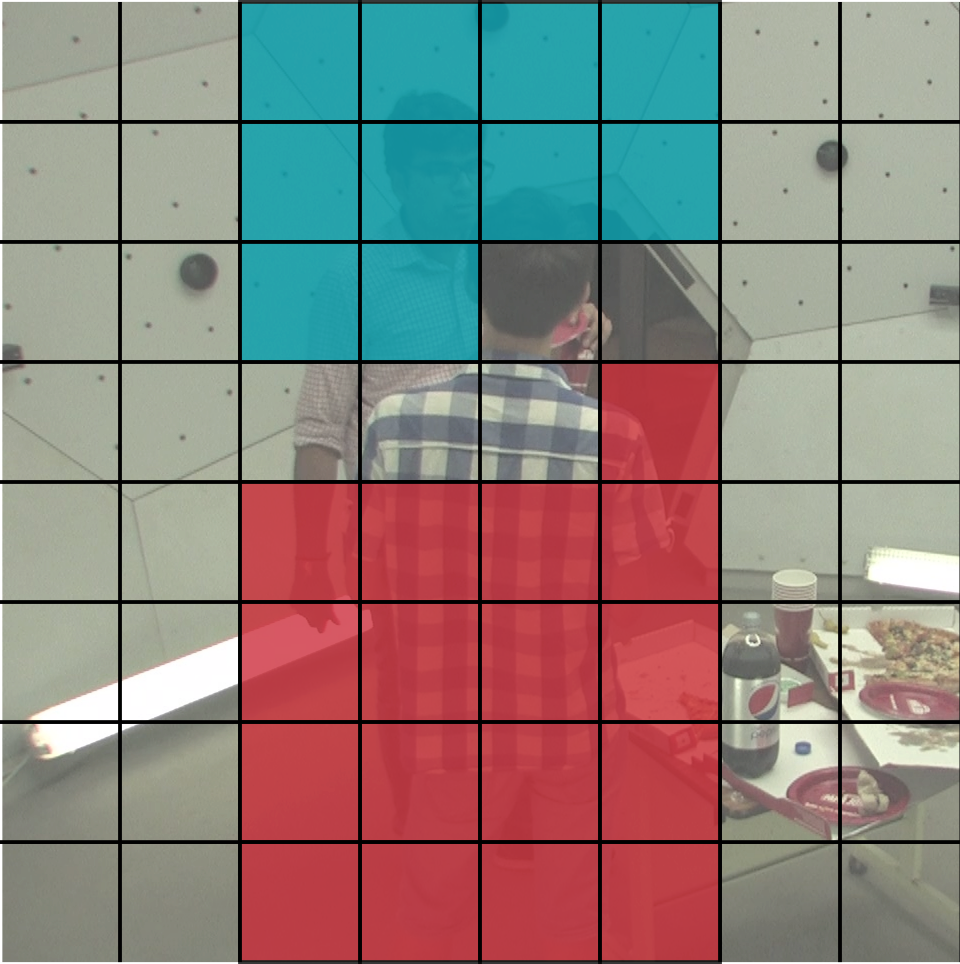}
        \caption{Selected Anchors $\mathcal{A}^{++}$ }\label{fig:A++}
    \end{subfigure}

\caption{\textbf{Pose-Aware Anchors Selection}. A grid of anchors $\mathcal{A}$ is first computed at all output 2D positions (Figure \ref{fig:paas_A}). An example of an anchor is depicted in yellow.  
The matched anchors $\mathcal{A}^+$ correspond to anchors with a sufficient PONO (depicted in red and blue in Figure \ref{fig:passA+} ).  
Nevertheless, some of these anchors are ambiguous (crossed anchors in  \ref{fig:passA+}) as they correspond to overlapping persons. They are filtered by the Pose-Aware Anchors Selection strategy to obtain the set of non-ambiguous positive readout anchors $\mathcal{A}^{++}$ depicted in Figure \ref{fig:A++} (best viewed in color).} 
\label{fig:paas}
\end{figure}

\subsection{Pose-Aware Anchor Selection}
\label{ss:pose-aware}

As illustrated in Figure \ref{fig:paas}, some of the positive anchors in $\mathcal{A}^+$ are not suited for the readout of consistent human poses. 
When several subjects overlap, these anchors may contain more than one person. This can lead to erroneous predicted bounding boxes and incorrect human poses. Consequently, at inference, precise readout locations are needed to determine the final bounding boxes and poses. To do so, the network should be trained to consider ambiguous anchors as negative. We define $\mathcal{A}^{++}$ to be the set of non ambiguous readout anchors. 

A way to filter $\mathcal{A}^{+}$ to get $\mathcal{A}^{++}$ is to threshold the product of the overlap between ground truth and predicted bounding boxes $\hat{O}_{i,j,a}$ and the PONO value $O_{i,j,a}$ , as it is done in \cite{chabot}. In other words, an anchor belongs to $\mathcal{A}^{++}$ if the box predicted by this anchor correctly fit its associated ground truth. For detection purpose this strategy may be sufficient to solve ambiguities, but for pose estimation such a filtering is too coarse.
Anchors in $\mathcal{A}^{++}$ must lead to the readout of valid and unambiguous human poses. To this end, we introduce a Pose-Aware Anchor Selection strategy based on 2D poses overlap. This overlap $\hat{O}^{2D}_{i,j,a}$ is defined as the mean of $\hat{O}^{2D}_{i,j,a,k}$ for all joints $k$ of the subject.

Thus, the Positive Readout Anchors Labels ($C_{i,j,a}$) are defined by : 

\begin{equation}
    C_{i,j,a}= \left\{
    \begin{array}{ll}
        1 & \mbox{if } O_{i,j,a}  \times \hat{O}^{2D}_{i,j,a}>0.5 \\
        0 & \mbox{otherwise}
    \end{array}
\right.
\label{equ:pral}
\end{equation}

The pixel-wise classification loss is then defined by: 

\begin{equation}
    \mathcal{L}_{cls}(i,j,a)=\mathcal{H}(C_{i,j,a},\hat{C}_{i,j,a})
\end{equation}

where $\mathcal{H}$ is the standard binary cross-entropy.

\subsection{Automatic weighting of losses with homoscedastic uncertainty}
\label{ss:alw}

A classical problem in training a multi-task network is to properly weight each task. 

Kendall \textit{et al.} \cite{kendall2018multi} propose a loss based on the homoscedastic uncertainty (\textit{i.e} independent of the input data) to weight multiple losses in a multi-task network.  
Another issue in single-shot multi-person pose estimation is the imbalance between subject’s sizes in the image. In real-world images, there are both a large range of distances to the camera and an imbalance in people sizes in the image. In Lapnet \cite{chabot}, anchor weights are learned to solve this problem for rigid object detection. 
In multi-person pose estimation tasks of PandaNet, uncertainties related to joints have to be managed. Indeed, as joints have different degrees of freedom, predictions associated to hips are more certain than predictions associated to hands for instance. Uncertainty also depends on people sizes in the image. A network is less precise for people at low resolution than for high resolution people. Furthermore, far from the camera people are more prone to occlusions than other people making the regressed coordinates associated to these people more uncertain. This is why we propose to learn joint specific regression weights for each predefined anchor and introduce the following loss functions: 
%\small{
\begin{dmath}
   \mathcal{L}_{cls} = \frac{\lambda_{cls}}{HWN_A}\sum_a \lambda^a_{cls}\sum_{i,j}\mathcal{L}_{cls}(i,j,a) + \log(\frac{1}{\lambda_{cls}}) + \frac{1}{N_A}\sum_a \log(\frac{1}{\lambda_{cls}^a})
\end{dmath}

\begin{dmath}
   \mathcal{L}_{loc} = \frac{\lambda_{loc}}{N^+}\sum_a \lambda^a_{loc}\sum_{i,j}\mathcal{L}_{loc}(i,j,a) +  \log(\frac{1}{\lambda_{loc}}) +   \frac{1}{N_A}\sum_a \log(\frac{1}{\lambda_{loc}^a}) 
\end{dmath}

\begin{dmath}
   \mathcal{L}_{2D} = \frac{\lambda_{2D}}{N_KN^+} \sum_{i,j,a,k}\lambda^{a,k}_{2D}\mathcal{L}_{2D}(i,j,a,k) + \log(\frac{1}{\lambda_{2D}}) +  \frac{1}{N_KN_A}\sum_ {a,k} \log(\frac{1}{\lambda_{2D}^{k,a}}) 
\end{dmath}

\begin{dmath}
   \mathcal{L}_{3D} = \frac{\lambda_{3D}}{N_KN^+} \sum_{i,j,a,k}\lambda^{a,k}_{3D}\mathcal{L}_{3D}(i,j,a,k) + + \log(\frac{1}{\lambda_{3D}})+ \frac{1}{N_KN_A}\sum_ {a,k} \log(\frac{1}{\lambda_{3D}^{k,a}})
\end{dmath}
%}

where  $\lambda_{cls}$, $ \lambda_{loc}$, $\lambda_{2D}$ and  $\lambda_{3D}$ are the task weights, $\lambda^a_{cls}$ and $\lambda^a_{loc}$ are the anchors weights and $\lambda^{a,k}_{2D}$ and $\lambda^{a,k}_{3D}$ are the anchor-joint regression weights. $N^+$ is the number of anchors in $A^+$. All weights $\lambda$ are trainable variables. All terms $\log(\frac{1}{\lambda})$ are regularisation terms that avoid all $\lambda$ to converge to $0$. The final total loss is :
\begin{dmath}
   \mathcal{L}_{total} = \mathcal{L}_{cls} + \mathcal{L}_{loc} + \mathcal{L}_{2D} + \mathcal{L}_{3D}
\end{dmath}

\section{Experimental Results}

PandaNet's performance is evaluated on three datasets: JTA  \cite{fabbri2018learning}, CMU-Panoptic \cite{Joo_2017_TPAMI} and MuPoTS-3D \cite{mehta2017single}.

\textbf{Evaluation Metrics: } To evaluate multi-person 3D pose approaches, we use two metrics. The first one is the Mean per Joint Position Error (MPJPE) that corresponds to the mean Euclidean distance between the ground truth and the prediction for all people and all joints. The second one is the 3DPCK which is the 3D extension of the Percentage of Correct Keypoints (PCK) metric used for 2D pose evaluation. A joint is considered correctly estimated if the error in its estimation is less than 150mm. If an annotated subject is not detected by an approach, we consider all of its joints to be incorrect in the 3DPCK metric. The human detection performance is evaluated with the Average Precision (AP) used in the PASCAL VOC challenge \cite{Everingham15}.

\textbf{Training Procedure:}  The method was implemented and tested with TensorFlow 1.12.  In all our experiments, the model is trained with mini-batches of 24 images. SGD optimiser is used with a momentum of 0.9,  an initial learning rate of 0.005 and a polynomial decay policy of 0.9. Random crops and random scales are used for data augmentation. Synchronized batch normalisation across GPU is used.  Darknet-53 \cite{redmon2018yolov3} is used as backbone. The number of anchors $N_A$ is set to 10 in all experiments.

\subsection{JTA dataset results}

JTA (Joint Track Auto) is a dataset for human pose estimation and tracking in urban environment. It was collected from the realistic video-game the Grand Theft Auto V and contains 512 HD videos of 30 seconds recorded at 30 fps.  The collected videos feature a vast number of different body poses, in several urban scenarios at varying illumination conditions and viewpoints. People perform different actions like walking, sitting, running, chatting, talking on the phone, drinking or smoking. Each image contains a number of people ranging between 0 and 60 with an average of 21 people. The distance from the camera ranges between 0.1 to 100 meters, resulting in pedestrian heights between 20 and 1100 pixels. No existing dataset  with  annotated  3D  poses  is  comparable  with  JTA  dataset in terms of number and sizes of people. An input image size of 928x576 is used.

\begin{table}
\centering
%\resizebox{0.7\linewidth}{!}{%
\begin{tabular}{l|ll}
Anchor Selection    & AP  & 3DPCK  \\ \hline
No                  & 84.1                   & 80.7                  \\
BB-Aware \cite{chabot}              & 85.1                    & 81.9              \\
Pose-Aware (Ours)    & \textbf{85.3}  & \textbf{83.2} 
\end{tabular}
%}

\caption{Influence of the Anchor Selection Strategy. All the models are trained with the Automatic Weighting of Losses.}
\label{tab:aw_jta}
\end{table}

\begin{table}
\centering
%\resizebox{\linewidth}{!}{%
\begin{tabular}{l|l|l|ll}
%$\lambda_{loc}$, $\lambda_{cls}$, $\lambda_{2D}$ , $\lambda_{3D}$    & $\lambda_{loc}^a$ , $\lambda_{cls}^a$ & $\lambda_{2D}^{a,k}$, $\lambda_{3D}^{a,k}$ & $AP^{box}$  & 3DPCK        \\ \hline
task & anchor & joint & AP & 3DPCK        \\ \hline
1            & 1          & 1                 & 21.7                & 15.8          \\
learned & 1           & 1                    & 84.1               & 80.8          \\
learned  & learned & 1                    & 85.2                 & 81.7          \\
learned  & learned & learned         &  \textbf{85.3} & \textbf{83.2}
\end{tabular}
%}

\caption{Influence of the Automatic Weighting of Losses. task, anchor and joint represent the type of trainable $\lambda$ weights. All the models are trained with the Pose-Aware Anchor Selection Strategy.}
\label{tab:alw_jta}

\end{table}

\subsubsection{Ablation Studies}

\textbf{Pose-Aware Anchor Selection strategy: } Table \ref{tab:aw_jta} results show the effectiveness of the Pose-Aware Anchor Selection. We compare three variants of PandaNet. The first variant (first line) is a model where no anchor selection strategy is used. It corresponds to a model where only the PONO overlap $O_{i,j,a}$ is considered in equation \ref{equ:pral}. Using the Bounding-box Aware Anchor Selection \cite{chabot} (second row), improves the model performance over this baseline. Box detection and 3D pose estimation take all benefit of this anchor selection strategy. Using the proposed Pose-Aware Anchor Selection (third row) maintain the AP value while improving the 3DPCK, showing its effectiveness for choosing better anchors for 3D pose estimation.

\textbf{Automatic Weighting of Losses:} The influence of Automatic Weighting of Losses is detailed in Table \ref{tab:alw_jta}. When the $\lambda$'s are all set to 1 (first line) and not trained, the model has poor performance on all tasks. Learning task-specific $\lambda_{loc}$, $\lambda_{cls}$, $\lambda_{2D}$  and $\lambda_{3D}$ (second row) allow the network to converge and to achieve good performances on all tasks.
Learning anchor weights $\lambda_{loc}^a$ and  $\lambda_{cls}^a$ (third row) improves detection and 3D pose estimation performances. The best results are obtained when all $\lambda$'s are learned, showing the importance of the proposed automatic weighting of losses.

\subsubsection{Comparison with prior work}

% \textbf{Comparison with a 3D human pose work:} 
The approach in \cite{benzinepr} is the only method that provides 3D results on the JTA dataset. We compare PandaNet with the best model in \cite{benzinepr} \textit{i.e} the model with multi-scale inference.

\begin{table}[t]
% \resizebox{\linewidth}{!}{
\centering
\begin{tabular}{|l|l|l|l|l|l|l|}
\hline
Dist. & \textless{}10 & 10-20 & 20-30 & 30-40 & \textgreater{}40 & All  \\ \hline
 \cite{benzinepr}  & 55.8           & 68.4                               & 57.8                              & 49.3                              & 41.7              & 43.9 \\
Ours    &\textbf{ 95.6}           & \textbf{93.7 }                              & \textbf{87.3 }                             & \textbf{80.5}                              & \textbf{71.2}              & \textbf{83.2} \\ \hline
\end{tabular}

% }

\caption{Distance wise 3DPCK on the JTA dataset. Distance are in meters.}
\label{tab:res_distance_joint_wise}
\end{table}

\begin{table*}[]
\centering
% \resizebox{\linewidth}{!}{
\begin{tabular}{|l|l|l|l|l|l|l|l|l|l|l|l|l|}
\hline
 Method                                   & head  & neck  & clavicles & shoulders & elbows & wrists & spines & hips.  & knees & ankles & all    \\ \hline
                                            \cite{benzinepr} & 41.1  & 44.6 &   44.9      & 33.8      & 27.2   & 19.0   & 74.4  & 73.9  & 25.7  & 8.9    & 43.9    \\ 
                                                Ours & \textbf{92.7}  & \textbf{99.1} &   \textbf{97.0}      & \textbf{78.4}      & \textbf{72.1}   & \textbf{60.1}   & \textbf{99.9}  & \textbf{87.8}  & \textbf{71.8}  & \textbf{58.0}    & \textbf{83.2 }   \\\hline
\end{tabular}
% }

\caption{Joint wise 3DPCK.}
\label{tab:res_jta_joint_wise}
\end{table*}

Table \ref{tab:res_distance_joint_wise} provides 3DPCK results according to the distance of people to the camera. PandaNet outperforms the model of Benzine \etal \cite{benzinepr} on all camera distances demonstrating the ability of PandaNet to properly process people at all scales. While our model achieve very good results for people close to the camera (less than 20m), it also correctly handles people who are further from the camera.

Table \ref{tab:res_jta_joint_wise} provides joint-wise results on the JTA dataset. PandaNet outperforms the model of Benzine \etal \cite{benzinepr} for all joints. In particular, it has no difficulties to estimate 3D coordinates for the joints that have the fewest degrees of freedom (head, neck, clavicles , hips and spines) with a 3DPCK for these joints greater than 92\%. PandaNet increases the 3DPCK for the shoulders by 44.6\% and for the elbows by 34.9\%. Terminal joints (wrists and ankles) are the most difficult joints with a 3DPCK of 60.1\% and 58.0\% for these joints against 19.0\% and 8.9\% for \cite{benzinepr}.

\subsection{CMU-Panoptic results}

CMU Panoptic \cite{Joo_2017_TPAMI} is a dataset containing images with several people performing different scenarios in a dome where many cameras are placed. Although it was acquired in a simple and controlled environment, this dataset is challenging because of complex interactions and difficult camera viewpoints. We evaluate PandaNet using the protocol used in \cite{zanfir2018monocular,zanfir2018deep, benzine2019deep}. The test set is composed of 9600 frames from HD cameras 16 and 30 and for 4 scenarios: Haggling, Mafia, Ultimatum, Pizza. The model is trained on the other 28 HD cameras of CMU Panoptic. An input image size of 512x320 is used on all Panoptic experiments.

On this dataset, PandaNet improves the results over the recent state of the art methods on all scenarios (Table \ref{tab:res_panoptic}). The average MPJPE is improved by 25.8mm compared to the best previous approach.  While the results on JTA prove the ability of the model to deal with realistic urban scenes with many people at low resolution, results on the Panoptic dataset show that the approach is effective to manage people overlaps and crops that frequently occur in this dataset.

\begin{table}[t]
\centering
%\resizebox{\textwidth}{!}
%\resizebox{\linewidth}{!}{%
\begin{tabular}{@{}llllll@{}}
\toprule
Method    & Haggling & Mafia & Ultimatum & Pizza & Mean  \\ \midrule
%  \cite{popa2017deep} & 217.9    & 187.3 & 193.6     & 221.3 & 203.4 \\
 \cite{zanfir2018monocular}   & 140.0    & 165.9 & 150.7     & 156.0 & 153.4 \\
 \cite{zanfir2018deep}   & 72.4     & 78.8  & 66.8      & 94.3  & 72.1  \\ 
 \cite{benzine2019deep}       & 70.1   & 66.6 & 55.6    & 78.4  & 68.5 \\ \midrule
Ours      & \textbf{40.6}    & \textbf{37.6} & \textbf{31.3}     & \textbf{55.8}  & \textbf{42.7} \\ \bottomrule
\end{tabular}
%}

\caption{MPJPE in mm on the Panoptic dataset.}
\label{tab:res_panoptic}
\end{table}

\subsection{MuPoTS-3D results}

\begin{table}[t]
\centering
%\resizebox{0.7\linewidth}{!}{

\begin{tabular}{@{}cc|c@{}}
\toprule
\multicolumn{2}{c|}{Method}                                    & 3DPCK \\ \midrule
\multicolumn{1}{c|}{\multirow{3}{*}{Two-Stage}}  & LCR-Net\cite{rogez2017lcr}   & 53.8  \\
\multicolumn{1}{c|}{}                             & LCR-Net++ \cite{rogez2019lcr} & 70.6  \\
\multicolumn{1}{c|}{}                             & Moon \etal \cite{moon2019camera}      & \textbf{81.8}  \\ \midrule
\multicolumn{1}{c|}{\multirow{3}{*}{Single-Shot}} & Mehta \etal \cite{mehta2017single}  & 66.0  \\
\multicolumn{1}{c|}{}                             & XNect \cite{mehta2019xnect}   & 70.4  \\
\multicolumn{1}{c|}{}                             &      PandaNet     & \textbf{72.0 } \\ \bottomrule
\end{tabular}

%}

\caption{3DPCK on the MuPoTS-3D dataset.}
\label{tab:mupots}
\end{table}

\begin{figure}[t] % "0.45" donne ici la largeur de l'image
\centering
\includegraphics[width=\linewidth]{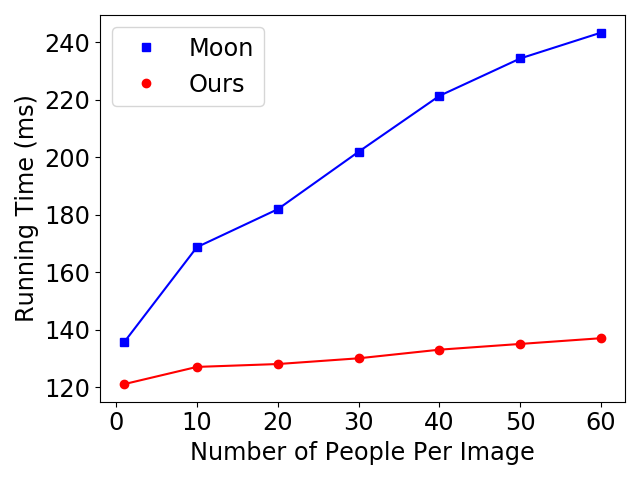}
        \caption{Running time comparison with the approach of Moon \etal \cite{moon2019multi} according to the number of people per image. Experiments are performed on a NVIDIA Titan X GPU. Images come from JTA Dataset \cite{fabbri2018learning}.   }\label{fig:RT}
\end{figure}

MuPoTS-3D  \cite{mehta2017single} is a dataset containing 20 sequences with ground truth 3D poses for up to three subjects. PandaNet is trained on the MuCo-3DHP dataset that is generated by compositing the existing MPI-INF-3DHP 3D single-person pose estimation dataset \cite{mehta2016monocular}, and on the COCO-dataset \cite{lin2014microsoft} to ensure better generalisation. Each
mini-batch consists of half MuCo-3DHP and half COCO
images. For COCO data, the loss value for the 3D regression task is set to zero. An input image size size of 512x512 is used and Subpixel convolutions are removed.

Table  \ref{tab:mupots} provides  3DPCK  results on this dataset. PandaNet achieve higher 3DPCK than previous single-shot approaches. It improves over an ORPM method \cite{mehta2017single} by 6\% and over XNect \cite{mehta2019xnect} by 1.6\%. XNect is composed of two different models. The first one predicts partial 2D and 3D pose encoding and the second one refines these encodings to get final full 3D poses. Consequently, the weaknesses of the first model (like joints occlusions and people crops) are compensated by the second one. We achieve better results with a single model without any refinement process. Compared to two-stage models, PandaNet achieves better results than LCR-Net \cite{rogez2017lcr} and LCR-Net++ \cite{rogez2019lcr}. Compared to the approach of Moon \etal \cite{moon2019multi},  PandaNet has a lower 3DPCK. This top-down approach uses an external two-stage object detector (Faster-R CNN \cite{ren2015faster}) to compute human bounding boxes and forward each cropped subject to a single-person 3D person approach \cite{sun2018integral}. Therefore, the computation complexity of their model  depends on the number of people in the image like illustrated in Figure \ref{fig:RT}. If the number of people is large, this approach scales badly.  On the contrary,  the proposed  single-shot model allows a nearly constant inference time regarding the number of people. The inference time of PandaNet is about 140ms for images with 60 people on a NVIDIA Titan X GPU.

\section{Conclusion}

PandaNet is a new anchor-based single-shot multi-person pose estimation model that efficiently handles scene with a large number of people, large variation in scale and people overlaps. This model predicts in a single-shot way people bounding boxes and their corresponding 2D and 3D pose. To properly manage people overlaps, we introduce a Pose-Aware Anchor Selection strategy that discards ambiguous anchors. Moreover, an automatic weighting has been provided for three main purposes. It balances task-specific losses, it compensates imbalance in people sizes and it manages uncertainty related to joints coordinates.

The experiments validate the proposed Anchor-based Multi-person Pose Regression framework and prove the importance of the Pose-Aware Anchor Selection strategy and of the Automatic Weighting. Furthermore, large-scale experiments, on JTA, CMU Panoptic, and MuPoTS-3D datasets demonstrate that PandaNet outperforms previous single-shot state of the art methods.

{\small
\bibliographystyle{ieee_fullname}
\bibliography{egbib}
}
% \end{multicols}

\end{document}